%% file: example_paper.tex
\documentclass{article}

\usepackage{microtype}
\usepackage{graphicx}
\usepackage{subcaption}
\usepackage{booktabs} 

\usepackage{hyperref}

\usepackage[preprint]{icml2026}

\usepackage{amsmath}
\usepackage{amssymb}
\usepackage{mathtools}
\usepackage{amsthm}

\usepackage{xspace}
\usepackage[table]{xcolor}
\usepackage{amssymb}
\usepackage{float}
\usepackage{multirow}

\newcommand{\ie}{\emph{i.e.}\xspace}
\newcommand{\eg}{\emph{e.g.}\xspace}
\newcommand{\methodnamefull}{Relational and Structural Consistency Network\xspace}
\newcommand{\methodname}{RSCN\xspace}

\usepackage[capitalize,noabbrev]{cleveref}

\theoremstyle{plain}

\theoremstyle{definition}

\theoremstyle{remark}

\usepackage[textsize=tiny]{todonotes}

\icmltitlerunning{Instance-Free Domain Adaptive Object Detection}

\begin{document}

\twocolumn[
  \icmltitle{Instance-Free Domain Adaptive Object Detection}



  \icmlsetsymbol{equal}{*}

  \begin{icmlauthorlist}
    \icmlauthor{Hengfu Yu}{yyy}
    \icmlauthor{Jinhong Deng}{yyy}
    \icmlauthor{Lixin Duan}{yyy}
    \icmlauthor{Wen Li}{yyy}
  \end{icmlauthorlist}

  \icmlaffiliation{yyy}{School of Computer Science and Engineering, University of Electronic Science and Technology of China, Chengdu, China}
    
  \icmlcorrespondingauthor{Wen Li}{liwenbnu@gmail.com}

  \icmlkeywords{Computer Vision, Domain Adaptation, Object Detection, ICML}

  \vskip 0.3in
]



\printAffiliationsAndNotice{}  

\input{sections/0_abstract}
\input{sections/1_introduction}
\input{sections/2_related_works}

\input{sections/3_method}
\input{sections/4_experiments_datasets}
\input{sections/4_experiments_method}
\input{sections/5_conclusion}


\newpage
\section*{Impact Statement}
The goal of this research is to advance the field of Domain Adaptive Object Detection by addressing the practical challenge of missing target instances. Our method, \methodname, reduces the reliance on acquiring rare positive samples, thereby promoting the broader applicability of object detection models. This efficiency is particularly valuable for applications with high social impact, such as healthcare analysis and ecological protection. We believe this work contributes to the development of more robust and deployable machine learning systems.

\bibliography{example_paper}
\bibliographystyle{icml2026}

\newpage
\input{sections/6_supplementary}



\end{document}

%% file: sections/0_abstract.tex
\begin{abstract}
While Domain Adaptive Object Detection (DAOD) has made significant strides, most methods rely on unlabeled target data that is assumed to contain sufficient foreground instances. However, in many practical scenarios (e.g., wildlife monitoring, lesion detection), collecting target domain data with objects of interest is prohibitively costly, whereas background-only data is abundant. This common practical constraint introduces a significant technical challenge: the difficulty of achieving domain alignment when target instances are unavailable, forcing adaptation to rely solely on the target background information. We formulate this challenge as the novel problem of Instance-Free Domain Adaptive Object Detection. To tackle this, we propose the \methodnamefull (\methodname) which pioneers an alignment strategy based on background feature prototypes while simultaneously encouraging consistency in the relationship between the source foreground features and the background features within each domain, enabling robust adaptation even without target instances. To facilitate research, we further curate three specialized benchmarks, including simulative auto-driving detection, wildlife detection, and lung nodule detection. Extensive experiments show that \methodname significantly outperforms existing DAOD methods across all three benchmarks in the instance-free scenario. The code and benchmarks will be released soon.
\end{abstract}

%% file: sections/1_introduction.tex
\begin{figure}[t]
  \centering
   \includegraphics[width=1.0\linewidth]{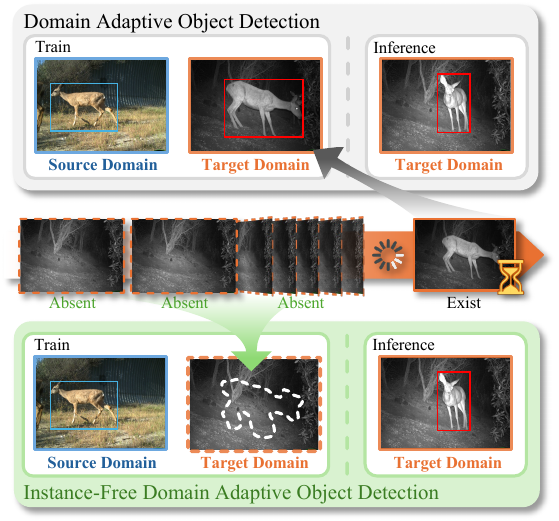}

   \caption{
   Illustration of Instance-Free DAOD. The orange timeline indicates that target-domain instances appear only after a long waiting period and require costly manual screening. Traditional DAOD (top) relies on target images containing foreground instances, whereas Instance-Free DAOD (bottom) enables transferring using only background-only target images.
   }
   \label{fig:1_intro_if_daod}
\end{figure}

\section{Introduction}
\label{sec:intro}

Driven by advances in deep learning\cite{vgg, resnet, transformer}, object detection\cite{rcnn, fasterrcnn, yolo, retinanet, detr, deformabledetr, dino} has achieved remarkable success. However, a well-known challenge is the significant performance degradation when models are applied to new, unseen domains. To address this, Domain Adaptive Object Detection (DAOD)\cite{daf, swda, divmatch, scda, htcn, everypixelmatters, megacda, sigma, umt, pt, at, ssal, mttrans, cmt, cat} has been proposed, where a detector is trained on fully labeled source-domain data and unlabeled target-domain data to mitigate the distributional shift between the two domains.

While this setting is reasonable when target-domain images containing foreground objects are readily available, this assumption often breaks down in real applications. In scenarios like wildlife monitoring\cite{cct} and wilderness search and rescue~\cite{rescue}, positive instances are inherently rare. In cross-device medical imaging, such as lung nodule detection\cite{luna16}, newly deployed scanners may not gather enough lesion-positive cases in the short term. In contrast, background-only target data, such as empty natural scenes or healthy patient scans, can be collected easily and at scale. Under this constraint, existing DAOD methods face a fundamental obstacle: they inherently rely on target-domain foreground instances to guide distribution alignment. When target foregrounds are absent, source foregrounds are at risk of misalignment with target backgrounds, and pseudo-labels collapse under false positives, thereby causing standard DAOD methods to fail. This motivates a critical question: \textit{Can we achieve domain adaptation using only background information from the target domain?}

Therefore, in this paper, we tackle this problem, which we term Instance-Free Domain Adaptive Object Detection (Instance-Free DAOD), as illustrated in \cref{fig:1_intro_if_daod}. The core challenge is to achieve effective cross-domain alignment using source-domain foreground and background information together with background-only target data.
We introduce \methodnamefull (\methodname), a prototype-based framework tailored for Instance-Free DAOD. \methodname aligns background prototypes across domains and regularizes their relative geometry with respect to source foreground prototypes while preserving the discriminative structure of the source feature space.
To facilitate further studies, we establish three new benchmarks: a simulation dataset based on CARLA\cite{carla} and two challenging real-world benchmarks for the detection of wildlife and pulmonary nodules. Experiments show that previous DAOD methods often fail under the Instance-Free scenario, whereas our proposed method effectively improves the cross-domain performance. We summarize the contributions of this work as follows:
\begin{itemize}
\item We formalize the Instance-Free Domain Adaptive Object Detection (Instance-Free DAOD) problem, where only background images are available in the target domain, and standard DAOD assumptions no longer hold.
\item We propose the \methodnamefull (\methodname) to specifically tackle the challenge of Instance-Free DAOD, enabling robust adaptation where target foregrounds are absent.
\item We construct three specialized benchmarks targeting the practical challenge of Instance-Free DAOD, spanning autonomous driving, wildlife monitoring, and medical imaging scenarios.
\item Extensive experiments on all three benchmarks show that existing DAOD methods often fail in the Instance-Free setting, whereas \methodname significantly improves cross-domain detection performance.
\end{itemize}

%% file: sections/2_related_works.tex
\section{Related Works}

\subsection{Object Detection}

Object detection is a fundamental task in computer vision that involves the localization and classification of object instances within an image. Deep Convolutional Neural Networks (CNNs)\cite{lenet} led this field to two dominant paradigms: two-stage and one-stage detectors. The two-stage approach is pioneered by the R-CNN family \cite{rcnn, fastrcnn, fasterrcnn, maskrcnn, cascadercnn} and solidified by Faster R-CNN \cite{fasterrcnn}, employs a coarse-to-fine strategy. It first generates a sparse set of candidate proposals and subsequently performs classification and bounding box refinement. One-stage detectors, such as YOLO\cite{yolo} and SSD\cite{ssd}, eschew the discrete proposal generation stage, instead formulating the detection problem as a direct regression and classification task over a dense spatial grid, enabling real-time inference. Recently, anchor-free methods, including FCOS\cite{fcos} and CenterNet\cite{centernet} obviated the need for hand-crafted anchor box configurations. DETR\cite{detr} leverages Transformers\cite{transformer} to cast object detection as an end-to-end set prediction problem. Deformable DETR\cite{deformabledetr} and DINO\cite{dino} have made subsequent refinements to this philosophy. These detectors generally assume that training and test data follow the same \textit{i.i.d.}\ distribution. However, this assumption often fails in practice, such as when transferring from synthetic to real imagery or across different environmental conditions. This leads to substantial domain shift\cite{theory} and degraded performance on the target data. 

\begin{figure*}[!t]
  \centering
   \includegraphics[width=0.95\linewidth]{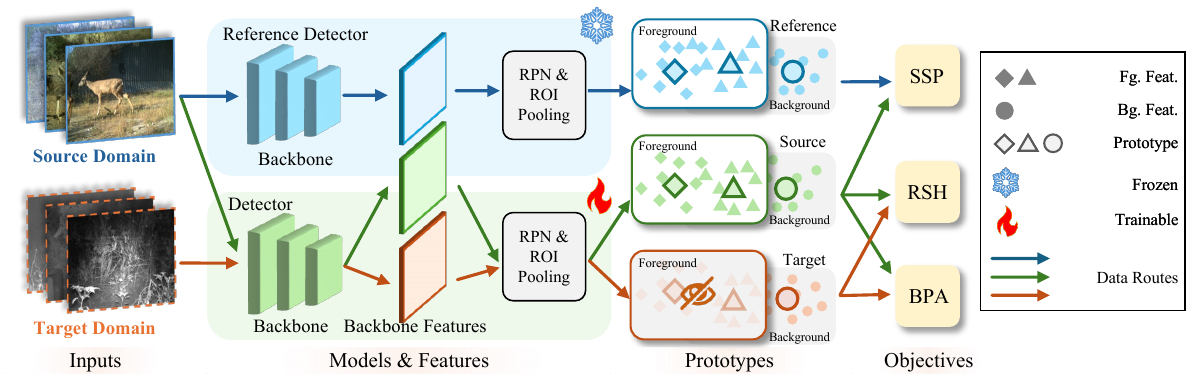}

   \caption{An overview of the proposed \methodname. For every batch, source-domain images with labeled foreground objects and target-domain images without foreground objects are fed to the detector. The background prototypes are aligned with the BPA objective in Eq.~\ref{eq:bpa}. RSH in Eq.~\ref{eq:rsh} keeps the relative geometric relationship for the source and target background prototypes to the shared foreground prototype anchors. A frozen reference detector is utilized to maintain the source-domain structure in order to avoid the feature collapse with SSP in Eq.~\ref{eq:ssp}.}
   \label{fig:2_framework}
\end{figure*}

\subsection{Domain Adaptive Object Detection}

To address the domain shift challenge outlined in the previous section, Domain Adaptive Object Detection (DAOD) has been extensively studied. Existing approaches can be broadly categorized into three paradigms: pixel-level, feature-level, and label-level adaptation.
Early approaches~\cite{htcn, umt} sought to mitigate the domain shift at the pixel level by employing the style transfer~\cite{cyclegan} technique.
Feature-level alignment serves as a dominant paradigm, typically achieving domain invariance via adversarial learning\cite{multi, divmatch, scda, everypixelmatters, vdd, mad}, inspired by DANN\cite{dann}. The pioneering DA Faster R-CNN\cite{daf} introduced domain classifiers with a gradient reversal layer (GRL) on image and instance levels to learn domain-invariant features. Subsequent works further refined this, including aligning multi-level backbone features\cite{htcn} with style transfer, focusing on local region alignment\cite{swda}, exploiting memory-guided attention mechanisms\cite{megacda}, and in a graph-matching way\cite{sigma, sigma++}.
Label-level self-training\cite{umt, pt, ssal, mttrans, cmt, cui2025debiased, he2025differential} has emerged as a powerful strategy more recently. These methods typically generate pseudo-labels for target domain images to guide a student detector, often within a mean-teacher~\cite{mt} framework (MT). AT\cite{at} enhances self-training with adversarial alignment. HT\cite{ht} introduces a Harmony Measure to assess pseudo-label quality. CAT\cite{cat} introduces an instance-level MixUp strategy that blends cropped instance patches across domains to mitigate pseudo-label bias caused by the class imbalance problem. While DAOD has advanced significantly, existing methods typically rely on the availability of target foregrounds. This assumption often fails in practice due to the extreme sparsity of foregrounds, in contrast to the ubiquity of background-only data.

%% file: sections/3_method.tex
\section{Method}

\subsection{Instance-Free DAOD Problem}

In the Unsupervised Domain Adaptive Object Detection framework, we are given a labeled source domain $\mathcal{D_S} = \{(x_i^s, y_i^s)\}_{i=1}^{n_s}$ drawn from a distribution $\mathcal{S}$. Here, $y_i^s$ denotes the ground-truth bounding boxes and class labels for image $x_i^s$, and the source foreground class set is $\mathcal{C_S}$. We are also given an unlabeled target domain $\mathcal{D_T} = \{x_i^t\}_{i=1}^{n_t}$ drawn from a target distribution $\mathcal{T}$. The standard Closed-Set setting, which we follow, assumes the label spaces are shared, \ie, $\mathcal{C_S} = \mathcal{C_T}$, and that $\mathcal{D_T}$ contains representative unlabeled foreground instances. The goal is to learn a detector $G$ using $\mathcal{D_S}$ and $\mathcal{D_T}$ that performs well on the target distribution $\mathcal{T}$. The learning objective can be written as:
\begin{align}
    &\theta^{*}
    = \arg\min_{\theta}\, \mathcal{L}(D_{\mathcal{S}}, D_{\mathcal{T}};\theta), \\
    \quad&\text{s.t.}\quad
    \begin{cases}
    D_{\mathcal{S}} \sim P_{\mathcal{S}}(X_{\mathcal{S}}^{\text{fg}}, X_{\mathcal{S}}^{\text{bg}}, Y_{\mathcal{S}}),\\
    D_{\mathcal{T}} \sim P_{\mathcal{T}}(X_{\mathcal{T}}^{\text{fg}}, X_{\mathcal{T}}^{\text{bg}}),
    \end{cases}
\end{align}
where $P_\mathcal{S}$ is the joint distribution over source-domain foregrounds, backgrounds, and labels, and $P_\mathcal{T}$ is the target-domain distribution over foregrounds and backgrounds. $\mathcal{L}$ represents the general DAOD objective, integrating the supervised detection loss on the labeled source data and additional adaptation regularization used to reduce the source-target discrepancy.

Instance-Free DAOD operates under a stricter information constraint on the target domain: during training, the available target data are instance-free images that contain only background. Let $D_\mathcal{T}^\text{free}$ denote such a dataset sampled from background-only marginal $P_\mathcal{T}^\text{bg}$ of the target domain. The learning problem becomes:
\begin{align}
    &\theta^{*}
    = \arg\min_{\theta}\, \mathcal{L}(D_{\mathcal{S}}, D_{\mathcal{T}}^\text{free};\theta), \\
    \quad&\text{s.t.}\quad
    \begin{cases}
    D_{\mathcal{S}} \sim P_{\mathcal{S}}(X_{\mathcal{S}}^{\text{fg}}, X_{\mathcal{S}}^{\text{bg}}, Y_{\mathcal{S}}),\\
    D_{\mathcal{T}}^\text{free} \sim P_{\mathcal{T}}^\text{bg}(X_{\mathcal{T}}^{\text{bg}}).
    \end{cases}
\end{align}

While the goal of generalizing to the target distribution remains unchanged, the learner no longer observes any target-domain foregrounds, receiving only background samples during training.
In summary, both standard DAOD and Instance-Free DAOD optimize the same target risk, and both have full access to labeled source-domain foregrounds. Yet, Instance-Free DAOD must learn to generalize to target-domain foregrounds while only observing background data from the target domain during training.

\begin{figure}[t]
  \centering
   \includegraphics[width=1.0\linewidth]{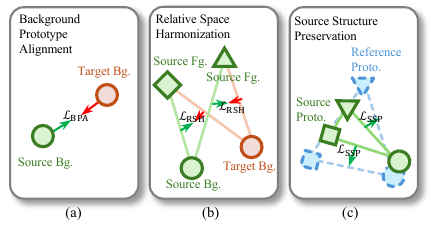}

   \caption{An overview of the proposed three constraints. (a) Background Prototype Alignment (BPA) minimizes the distance between the source and target background features. (b) Relative Space Harmonization (RSH) aligns the relationship between the source-domain foreground feature and background features from the two domains. (c) Source Structure Preservation (SSP) intends to maintain the feature relative structure in the source domain.}
   \label{fig:new_losses}
\end{figure}

\subsection{\methodnamefull}

Instance-Free DAOD poses a fundamental challenge for conventional domain-adaptive detectors: without any target-domain foreground instances, one cannot directly align object-level representations across domains.
While aligning background features mitigates domain discrepancy to some extent, it remains insufficient for effective adaptation. Despite the absence of target-domain foreground information, we explore cross-domain foreground-background relationships to further bridge the domain gap.


In this work, we design \methodnamefull (\methodname), whose framework is shown in \cref{fig:2_framework}. Our method introduces three complementary constraints, illustrated in \cref{fig:new_losses}: \emph{Background Prototype Alignment} (BPA) explicitly aligns class-agnostic background prototypes across domains; \emph{Relative Space Harmonization} (RSH) enforces consistency in the relative geometry between foreground and background prototypes across domains; and \emph{Source Structure Preservation} (SSP) regularizes the source feature space to maintain a discriminative object-centric structure while adapting. Together, these constraints harmonize the target background features with the source foreground–background structure, effectively mitigating the domain shift under instance-free supervision.

To implement these constraints, we construct class-wise prototypes for each input image to obtain robust and compact representations. Concretely, we extract average-pooled instance-level proposal features from the RPN of the Faster R-CNN detector $G$. For source-domain images in a batch, we average proposals according to their ground-truth categories to form foreground prototypes $\{p^s_c\}_{c \in C(x^s)}$, $p^s_c \in \mathbb{R}^d$, together with a source background prototype $p^s_{\text{bg}} \in \mathbb{R}^d$. Here $C(x^s)$ denotes the set of foreground class indices present in $x^s$, $\text{bg}$ is the index of the background class, and $d$ is the feature dimension. For target-domain images, which contain no foreground instances, we simply average all proposals to obtain a background prototype $p^t_{\text{bg}}$. These prototypes serve as basic units for methods in the following.

\noindent \textbf{Background Prototype Alignment (BPA).}
In Instance-Free DAOD, the only direct feature alignment is between the background class in the source and target domains. We propose to achieve this through Background Prototype Alignment (BPA). We introduce a background prototype domain discriminator $D_{\text{bg}}$, which is a three-layer MLP. Before inputting the background prototypes, a Gradient Reversal Layer (GRL) is applied. The discriminator $D_\text{bg}$ is trained to distinguish the domain label of the prototypes, while the feature extractor of detector $G$ is trained to produce features that $D_\text{bg}$ cannot distinguish. This adversarial process is optimized via a binary cross-entropy (BCE) loss:
\begin{equation}
\label{eq:bpa}
\begin{aligned}
    \mathcal{L}_\text{BPA} = &-\mathbb{E}_{x^s \sim \mathcal{D_S}}[\log D_\text{bg}(p_\text{bg}^s)] \\
    &- \mathbb{E}_{x^t\sim \mathcal{D}_\mathcal{T}^\text{free}}[\log(1-D_\text{bg}(p_\text{bg}^t))].
\end{aligned}
\end{equation}
The discriminator $D_\text{bg}$ minimized this loss, while the feature extractor maximizes it via the GRL.

\noindent \textbf{Relative Space Harmonization (RSH).}
One cannot directly align source foreground features with the target background features. However, the source foreground can still provide supervision through its relative geometry to background features. Once source and target backgrounds are aligned, their geometric relationships with respect to the shared source foreground anchors should also match. Relative Space Harmonization (RSH) enforces this consistency.

RSH utilizes the source foreground prototypes $\{p_c^s\}$ as anchors to aid and stabilize the alignment of $p_\text{bg}^s$ and $p_\text{bg}^t$.
We define this relationship from background to foreground anchors by computing the prototype vector difference. The source-domain relative vector is computed as $d_c^s = N(p_c^s - p_\text{bg}^s)$ and the cross-domain relative vector is computed as $d_c^t = N(p_c^s - p_\text{bg}^t)$. $N(\cdot)$ is the L2-normalization function. Then we propose the RSH loss, encouraging the geometric consistency by minimizing the distance between these two relative vectors:
\begin{equation}
\label{eq:rsh}
    \mathcal{L}_\text{RSH} = \sum_{c \in C(x^s)}
    {\|
        d_c^s - d_c^t
    \|}_1,
\end{equation}
where $C(x^s)$ is the set of foreground class indices present in $x^s$. $\mathcal{L}_\text{RSH}$ acts as a geometric consistency constraint, triangulating the positions of $p_\text{bg}^s$ and $p_\text{bg}^t$ from multiple viewpoints ($p_c^s$) to ensure the $\mathcal{L}_\text{BPA}$-driven alignment is robust and semantically meaningful.

\noindent \textbf{Source Structure Preservation (SSP).}
\label{sec:loss_diff}
While RSH effectively aligns the relative geometry across domains, it does not explicitly constrain the internal separability of the source domain. Without such regularization, the feature space remains vulnerable to structural degradation during adaptation~\cite{collapse, collapse2}, potentially leading to feature collapse where categories become indistinguishable. To address this, we introduce Source Structure Preservation (SSP) to regularize the intrinsic discriminative structure of the source features.

Here, we utilize the easily fetched source-only features trained only on the source-domain detection losses, as they can already provide a stable structural supervision. In each training iteration, we have the prototype set $P_S = \{p_c^s\}_{c \in C(x^s)} \cup \{p_\text{bg}^s\}$ from $G$, and suppose the source-only prototype set $P_R = \{p_c^{s'}\}_{c \in C(x^s)} \cup \{p_\text{bg}^{s'}\}$. Then we compute their respective inter-class cosine similarity matrices, denoted as $M_S$ and $M_R$. $\mathcal{L}_\text{SSP}$ is defined as the difference between these two structural relationship matrices:
\begin{equation}
    \label{eq:ssp}
    \mathcal{L}_\text{SSP} = 
    \sum_{i, j \in C(x^s) \cup \{\text{bg}\}, i \neq j}
    |(M_S)_{ij} - (M_R)_{ij}|.
\end{equation}

This preserves the discriminative structure learned in the source domain and prevents the feature space from collapsing during adaptation.

\subsection{Implementation of \methodname}
\noindent \textbf{Framework.} The framework details and the overall objective for the proposed \methodname is shown in \cref{fig:2_framework}.
During training, a reference detector $G_R$ with the same architecture as $G$ is introduced to provide the source structure supervision. In each batch, both networks receive the same source images. And the target images are only input into the detector $G$. Prototypes generated by $G$ from both the source and target domains are utilized to compute the BPA and RSH objectives. The prototypes generated from the reference detector $G_R$ are utilized to provide the SSP supervision signals. Note that the gradients from $\mathcal{L}_\text{SSP}$ are only propagated back to $G$. During inference, only the trained cross-domain detector $G$ is used for detection.

\noindent \textbf{Overall Objective.} The objective for the reference detector $G_R$ consists solely of the source detection loss:
\begin{equation}
    \mathcal{L}_{G_R} = \mathcal{L}_\text{det}(G_R(x^s), y^s),
\end{equation}
where $\mathcal{L}_\text{det}$ denotes the source-only detector loss.

The total objective for the cross-domain detector $G$ is the sum of the supervised detection loss and the three proposed constraints:
\begin{equation}
    \mathcal{L}_G = \mathcal{L}_\text{det}(G(x^s), y^s) + 
    \mathcal{L}_\text{BPA} + 
    \mathcal{L}_\text{RSH} +
    \mathcal{L}_\text{SSP}.
    \label{eq:total_loss}
\end{equation}

%% file: sections/4_experiments_datasets.tex
\begin{figure*}[]
  \centering
   \includegraphics[width=0.95\linewidth]{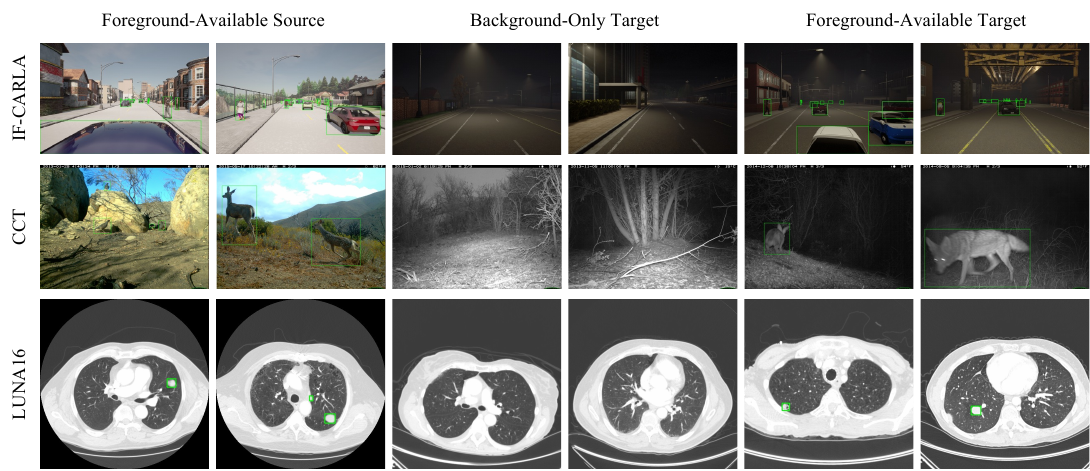}

   \caption{Examples of the three benchmarks. The IF-CARLA benchmark illustrates a domain shift from daytime to nighttime driving scenes. The IF-CCT benchmark captures a modality shift between visible-light imagery and infrared illumination. The IF-LUNA16 benchmark represents a cross-device transfer scenario, where differences in image layout, noise characteristics, and CT reconstruction protocols create inter-device domain discrepancies. Objects are marked with green bounding boxes.}
   \label{fig:3_benchmark}
\end{figure*}

\section{Experiment}

In this section, we first introduce three newly established benchmarks for Instance-Free DAOD to encourage further research. Based on the benchmarks, we compare our \methodname with the related DAOD approaches. Beyond performance comparison, we further provide more analysis, including the ablation study, the performance curve analysis, and the mechanism analysis of the proposed Relative Space Harmonization (RSH) and Source Structure Preservation (SSP). 

\subsection{Instance-Free DAOD Benchmarks}
\label{sec:benchmarks}
We propose three dedicated benchmarks to facilitate research in the challenging instance-free scenario. \cref{fig:3_benchmark} illustrates sample images, while detailed statistics information is available in the appendix.

\noindent \textbf{Simulative Auto-Driving Detection.} 
We propose the IF-CARLA dataset, constructed using the CARLA simulator \cite{carla}, to model the day-to-night domain shift in a strictly controlled environment. Spanning 4,430 viewpoints across 7 maps, the dataset comprises 13,290 images generated via a paired triplet strategy. For each viewpoint, we strictly control the camera pose to render a day-exist image (daytime with objects), a night-exist counterpart, and a night-free image (nighttime and background only). For experiments, we partition by location. The source is trained on 2,006 day-exist images with vehicles and pedestrians, while the target domain utilizes 1,820 night-free images for training and 604 night-exist images for validation.


\noindent \textbf{Wildlife Detection.} 
Acquiring sufficient positive samples (images containing animals) in the wild is a time-intensive process that can span multiple seasons. Also, as mentioned in \cite{cct}, the cameras are prone to false triggers from non-animal sources like wind or heat, generating a massive volume of empty frames. This underscores the necessity to bridge the domain gaps without waiting for costly positive sample collection. The CCT dataset~\cite{cct} is a benchmark for animal detection and classification, consisting of images collected from camera traps placed in the wild. There are two camera modalities in CCT. One consists of images captured in visible light, when ambient light is sufficient. Another consists of images captured using infrared (IR) illumination, when ambient light is insufficient.

In this work, we use CCT to model the pressing challenges in cross-domain wildlife monitoring. We construct the source-domain training set with 2,872 valid visible light images with animals. For the target domain, 2,678 IR instance-free images form the target domain training set, and 534 IR images with animals form the target domain validation set, named IF-CCT benchmark. We detect 5 animal classes: rabbit, deer, coyote, bobcat, and bird.

\noindent \textbf{Lung Nodule Detection.} 
In clinical practice, a number of labeled nodule cases are often available for an established CT scanner, whereas a newly deployed scanner typically lacks labeled positive samples. Acquiring sufficient positive data on the new device is slow and costly. Pulmonary nodules are relatively rare in the screening population, and even in nodule-positive scans, nodules occupy only a small fraction of slices, while most slices are nodule-free. This naturally aligns with the Instance-Free DAOD scenario. LUNA16 dataset~\cite{luna16} is a public benchmark in medical image analysis for lung nodule detection. It is a standardized subset of the LIDC-IDRI database~\cite{lidc-ldri}, which consists of numerous chest CT scans with nodule locations annotated by multiple radiologists. The LUNA16 dataset originates from CT scanners of different manufacturers, such as GE and Philips. Due to different scanners, reconstruction algorithms, and scanning parameters like layer thickness and X-ray dosage, the resulting images exhibit different layouts, textures, noise levels, and image contrast. 

We use 2,008 CT slices containing lung nodules from GE scanners as the source-domain training set. For the target domain, we use 865 nodule-free CT slices from Philips scanners as the target domain training set, and 231 Philips slices containing nodules as the target domain validation set, named the IF-LUNA16 benchmark.

%% file: sections/4_experiments_method.tex
\begin{table}[]
    \caption{Performance comparison on the IF-CARLA benchmark. mAP shows the average performance across all classes on AP@50. Gain shows the performance improvement or the degradation from the source-only model performance.}
    \label{tab:carla}
    \setlength{\tabcolsep}{2.8pt}
    \centering
    \begin{tabular}{lccccc}
    
    \hline
    Method      & Type  & Person   & Vehicle   & mAP       &  Gain   \\  \hline
    \rowcolor[HTML]{F5F5F5} 
    Source Only &       & 36.8     & 52.6      & 44.7      & -       \\
    CycleGAN    & Img. Trans.      
                        & 37.4     & 61.1      & 49.3      & {\color[HTML]{3531FF} +4.6}  \\
    \rowcolor[HTML]{F5F5F5} 
    DAF         &       & 37.8     & 57.3      & 47.6      & {\color[HTML]{3531FF} +2.9}  \\
    \rowcolor[HTML]{F5F5F5} 
    HTCN        &       & 37.7     & 61.6      & 49.7      & {\color[HTML]{3531FF} +5.0}  \\
    \rowcolor[HTML]{F5F5F5} 
    MAD         & \multirow{-3}{*}{Feat. Align.}
                        & 30.9     & 55.3      & 43.1      & {\color[HTML]{FE0000} -1.6}  \\
    AT          & \multirow{2}{*}{Self-Train.}
                        & 11.8     & 31.1      & 21.5      & {\color[HTML]{FE0000} -23.2}  \\
    CAT         &       & 29.5     & 46.5      & 38.0      & {\color[HTML]{FE0000} -6.7}   \\ \hline
    \rowcolor[HTML]{EFEFEF} 
    
    \textbf{\methodname} & & 41.2 & 68.4 & 54.8 & {\color[HTML]{3531FF} \textbf{+10.1}} \\ \hline
    \end{tabular}
\end{table}

\begin{table*}[]
    \caption{Performance comparison on the IF-CCT benchmark. mAP shows the average performance on AP@50 across all classes. Gain shows the performance improvement or the degradation from the source-only model performance.}
    \centering
    \setlength{\tabcolsep}{8pt}
    \label{tab:cct}
    \begin{tabular}{lcccccccc}
    \hline
    Method      & Type    & Rabbit & Deer  & Coyote & Bobcat & Bird  & mAP   & Gain         \\ \hline
    \rowcolor[HTML]{F5F5F5} 
    Source Only & & 40.7  & 46.7 & 56.6  & 36.2  & 32.3 & 42.5 & -                          \\
    CycleGAN    & Image Trans.
                & 24.1  & 57.3 & 56.4  & 42.0  & 33.1 & 42.6 & {\color[HTML]{3531FF} +0.1}  \\
    \rowcolor[HTML]{F5F5F5} 
    DAF         & 
                & 37.2  & 54.0 & 52.8  & 42.5  & 30.4 & 43.4 & {\color[HTML]{3531FF} +0.9}  \\
    \rowcolor[HTML]{F5F5F5} 
    HTCN        & 
                & 24.0  & 61.7 & 49.1  & 45.8  & 27.4 & 41.6 & {\color[HTML]{FE0000} -0.9}  \\
    \rowcolor[HTML]{F5F5F5} 
    MAD         & \multirow{-3}{*}{Feat. Align.}
                & 30.2  & 39.7 & 43.6  & 29.2  & 26.0 & 33.7 & {\color[HTML]{FE0000} -8.8} \\
    AT          & \multirow{2}{*}{Self-Train.} 
                & 1.6   & 15.7 & 9.8   & 10.9  & 1.9  & 8.0  & {\color[HTML]{FE0000} -34.5} \\
    CAT         & 
                & 21.2  & 18.7 & 37.1  & 30.1  & 2.9  & 22.0 & {\color[HTML]{FE0000} -20.5} \\ \hline
    \rowcolor[HTML]{EFEFEF} 
    \textbf{\methodname} & & 
                51.1  & 57.9 & 60.1  & 44.2  & 33.8 & \text{49.4} & {\color[HTML]{3531FF} \textbf{+6.9}}  \\ \hline
    \end{tabular}
\end{table*}

\begin{table}[!ht]
    \caption{Performance comparison on the IF-LUNA16 benchmark. mAP shows the performance on AP@50 of the nodule category. Gain shows the performance improvement or the degradation from the source-only model performance.}
    \label{tab:luna16}
    \centering
    \setlength{\tabcolsep}{8pt}
    \begin{tabular}{lccc}
    \hline
    Methods      & Type  & mAP   & Gain                           \\ \hline
    \rowcolor[HTML]{F5F5F5} 
    Source Only  &       & 39.1  & -                              \\
    CycleGAN     & Image Trans.
                         & 38.9  & {\color[HTML]{FE0000} -0.2}    \\
    \rowcolor[HTML]{F5F5F5} 
    DAF          &       & 38.1  & {\color[HTML]{FE0000} -1.0}    \\
    \rowcolor[HTML]{F5F5F5} 
    HTCN         &       & 32.0  & {\color[HTML]{FE0000} -7.1}    \\
    \rowcolor[HTML]{F5F5F5} 
    MAD          & \multirow{-3}{*}{Feat. Align.}
                         & 33.5  & {\color[HTML]{FE0000} -5.6}    \\
    AT           & \multirow{2}{*}{Self-Train.} 
                         & 8.9   & {\color[HTML]{FE0000}  -30.2}  \\
    CAT          &       & 10.4  & {\color[HTML]{FE0000} -28.7}   \\ \hline
    \rowcolor[HTML]{EFEFEF} 
    \textbf{\methodname} &
                        & 45.5  & {\color[HTML]{3531FF} \textbf{+6.4}} \\ \hline
    \end{tabular}
\end{table}

\begin{table}[]
    \caption{The ablation study for the three parts of the \methodname. By combining all three parts of \methodname, our method reaches a 10.1\% mAP gain among the source-only model.}
    \centering
    \label{tab:ablation}
    \begin{tabular}{l|ccc|cc}
    \hline
                        & BPA          & RSH          & SSP          & mAP   & Gain   \\ \hline
    Source Only         &              &              &              & 44.7  & -      \\ \hline
                        & $\checkmark$ &              &              & 48.9  & +4.2  \\
                        & $\checkmark$ & $\checkmark$ &              & 50.3  & +5.6  \\
                        & $\checkmark$ &              & $\checkmark$ & 52.1  & +7.4  \\ \hline
    \rowcolor[HTML]{EFEFEF} 
    \textbf{\methodname} & $\checkmark$ & $\checkmark$ & $\checkmark$ & \textbf{54.8} &  \textbf{+10.1} \\ \hline
    \end{tabular}
\end{table}


\subsection{Implementation Details}

Following the previous works\cite{daf, at, mad, cat} on DAOD, we implement our method using the Faster R-CNN detector with the Detectron2 framework. We adopt a ResNet-50 backbone pre-trained on ImageNet for all experiments. For training, we convert all three of the benchmarks into the PASCAL VOC format. All experiments are conducted on two NVIDIA RTX 3090 GPUs. The performance metrics used in our experiments are AP@50. More implementation details are provided in the appendix.

\subsection{Performance Comparison}
In this work, we compare the detection performance of the proposed \methodname on the three novel benchmarks mentioned, using six representative DAOD methods across the three transfer types.
Image-to-image style transfer~\cite{cyclegan} is a widely used technique in DAOD and serves as one of our baselines. DA Faster RCNN~\cite{daf} (we write it as DAF in the tables) is a classical DAOD method, aligning the image- and instance-level features between two domains. HTCN~\cite{htcn} is a representative method aligning multi-layer backbone features with the distribution enhancement by style transfer. MAD~\cite{mad} proposes a multi-view adversarial learning framework to mine and eliminate non-causal factors hidden in the features, thereby extracting purer domain-invariant representations. AT~\cite{at} seamlessly integrates pseudo-label supervision with backbone feature alignment within a mean-teacher architecture. CAT~\cite{cat} employs an instance-level MixUp strategy to blend cropped patches across domains. These methods are built on the Faster R-CNN detector and serve as typical examples of the existing DAOD methods.

\noindent \textbf{Simulative Auto-Driving Detection.} Performance comparison on IF-CARLA is shown in \cref{tab:carla}. 
Regarding existing paradigms, image-level translation and its combination with backbone feature alignment (HTCN) yield suboptimal improvements. Adversarial alignment methods such as DAF and MAD fail to effectively bridge the domain gap, as the absence of target foregrounds causes the discriminator to confuse source objects with target backgrounds, thereby degrading the discriminability of the detectors. The teacher-student frameworks suffer from a performance collapse, as erroneous pseudo-labels on background-only target images cause the student to learn incorrect foreground representations.
The \methodname guarantees the transfer safety and effectiveness in Instance-Free DAOD, and gains an improvement of 10.1\% mAP over the source-only performance, outperforming existing methods.

\noindent \textbf{Wildlife Detection.} In the IF-CCT benchmark, shown in \cref{tab:cct}. The detector transfers from the visible light mode to the infrared illumination mode for wildlife detection, which is a challenging real-world transfer scenario. Compared to synthetic samples, the transfer difficulty increases in the wildlife detection scenario, due to the diversity of detection samples and the inconsistency in overall image quality. While other methods fail to transfer effectively or suffer from significant performance degradation, our approach achieves a stable performance improvement of 6.9\% mAP.

\noindent \textbf{Lung Nodule Detection.} The IF-LUNA16 benchmark models a challenging medical lesion transfer detection scenario. Unlike salient objects in natural scenes, lung nodules are inherently subtle, exhibiting low contrast and high similarity to healthy tissue. Furthermore, significant cross-device layout discrepancies limit the efficacy of image-level transfer.
Despite this, as shown in \cref{tab:luna16}, our method also performs an effective transfer of a 6.4\% mAP gain.

\begin{figure}[t]
  \centering
   \includegraphics[width=0.95\linewidth]{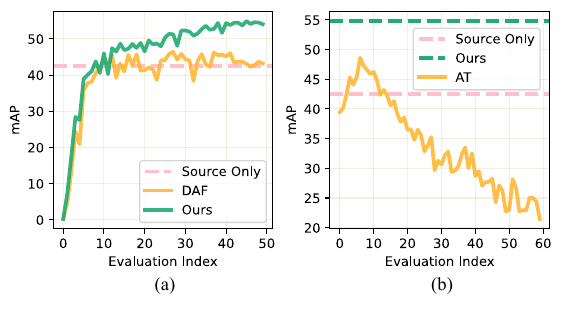}

   \caption{The performance comparison between \methodname and two comparison methods on IF-CARLA. Our method achieves better performance than (a) feature-alignment-based DAF~\cite{daf} and (b) self-training-based AT~\cite{at} baselines.
   }
   \label{fig:4_performance_drop}
\end{figure}

\begin{figure}[t]
  \centering
   \includegraphics[width=1.0\linewidth]{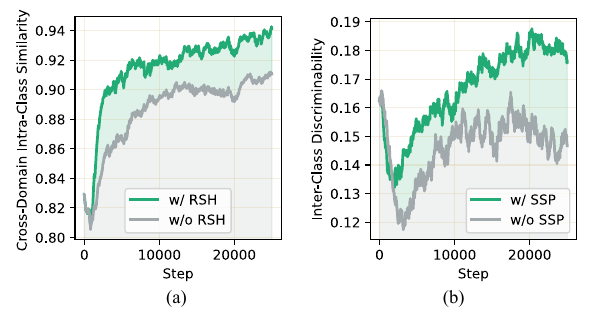}

   \caption{
  Mechanism analysis for RSH and SSP on IF-CARLA.
  (a) measures the average cosine similarity between source and target features of the same category. The results show that RSH steadily improves the alignment.
  (b) quantifies the separability among source class prototypes. SSP mitigates the feature collapse, safeguarding the intrinsic semantic structure throughout training.
}
   \label{fig:5_classwise_sim}
\end{figure}

\subsection{Further Analysis}

\noindent \textbf{Ablation Study.}
We perform the ablation study on the three parts of the Background Prototype Alignment (BPA), Relative Space Harmonization (RSH), and Source Structure Preservation (SSP) on the IF-CARLA dataset. As shown in \cref{tab:ablation}, the BPA gives an effective improvement of 4.2\% mAP, and this provides a foundation for the subsequent RSH and SSP. RSH introduces the extra relative structure supervision from the existing source-domain foreground features. The performance is improved to 5.6\% mAP. Just adding the RSH, the alignment is vulnerable to the feature degradation. Thus, the SSP is demanded. While the SSP itself can provide an improvement of 7.4\% in mAP with the BPA, it can further improve the performance with the final 10.1\% mAP together with the BPA and the RSH. By employing all parts of the \methodname, it provides an effective transfer in Instance-Free DAOD Scenarios.

\noindent \textbf{Performance Compared Across Training Iterations.}
We analyze the domain alignment and teacher-student self-training by examining their performance curves on IF-CARLA in \cref{fig:4_performance_drop}. Methods like DAF attempt to align features across domains, and the instance-level alignment could largely improve the transfer performance when target foreground instances exist. However, in the Instance-Free DAOD scenario, this leads to an erroneous alignment between source-domain foreground features and target-domain background features. This misalignment confuses the discriminative ability of the model. As shown in \cref{fig:4_performance_drop}(a), this causes the performance to fluctuate around the source-only baseline rather than achieving a stable improvement. The self-training methods like AT rely on pseudo-labels generated from target data. However, the teacher model generates false pseudo-labels on background-only images. This results in a rapid negative optimization and a catastrophic performance drop, as visualized in \cref{fig:4_performance_drop}(b).


\noindent \textbf{Mechanism Analysis on RSH and SSP.} We analyze the mechanism of RSH and SSP on IF-CARLA. 
First, to evaluate the contribution of RSH for the alignment process, we measure the Cross-Domain Intra-Class Similarity by calculating the average cosine similarity of each class across the source training and target validation sets. As shown in \cref{fig:5_classwise_sim}(a), RSH leads to a clear increase in similarity scores. This indicates that RSH effectively pulls the two domains closer.
Second, as discussed in \cref{sec:loss_diff}, we introduce SSP to mitigate the feature structural degradation during adaptation. We monitor the Inter-Class Discriminability on the source domain, defined as $1 - \frac{1}{N(N-1)}\sum_{i\neq j}^N \cos (p_i, p_j)$ where $N$ is the number of categories. As illustrated in \cref{fig:5_classwise_sim}(b), the integration of SSP maintains a higher score, demonstrating its effectiveness in safeguarding the intrinsic discriminative structure against collapse.

\noindent \textbf{Visualizations.} Due to the page limit, we provide visualizations of detection results in the appendix. 

%% file: sections/5_conclusion.tex
\section{Conclusion}

In this work, we introduce Instance-Free Domain Adaptive Object Detection, which is a practical yet previously overlooked scenario where only background-only images are available in the target domain during training. To address the fundamental challenge of training detectors that generalize to target-domain foregrounds under this constraint, we proposed the \methodnamefull (\methodname), which leverages the feature prototypes together with three complementary objectives: Background Prototype Alignment (BPA), Relative Space Harmonization (RSH), and Source Structure Preservation (SSP). We further established three dedicated benchmarks, IF-CARLA, IF-CCT, and IF-LUNA16, to study this problem across driving, wildlife monitoring, and medical imaging. Extensive experiments demonstrate that \methodname consistently gains improvements over the performance of the source-only model, outperforming representative DAOD methods, and show the effectiveness of the designed constraints. We hope that the Instance-Free DAOD formulation and the released benchmarks will foster future research on robust, deployment-ready transfer detectors and thereby broaden the practical applicability of domain adaptive object detection.

%% file: sections/6_supplementary.tex
\appendix

\section*{Appendix}
In this appendix, we provide additional details and further analysis for the main paper. The content is organized as follows:
\begin{itemize}
    \item \textbf{\cref{sec:supp_more_implementation_details}} elaborates on the implementation details, including the offline pre-computation strategy for the reference detector and specific training settings for each benchmark.
    \item \textbf{\cref{sec:supp_more_benchmark_information}} provides comprehensive information regarding the three proposed benchmarks (IF-CARLA, IF-CCT, and IF-LUNA16), including sample visualizations, detailed statistics, and data processing protocols.
    \item \textbf{\cref{sec:supp_detection_results_visualization}} presents extensive results, visually comparing the performance of our \methodname with the source-only baseline to demonstrate its robustness against domain shifts.
    \item \textbf{\cref{sec:supp_future_work_discussion}} discusses the potential future research direction to address the Instance-Free DAOD challenge.
\end{itemize}

\section{More Implementation Details}
\label{sec:supp_more_implementation_details}
In our \methodname framework, the reference detector $G_R$ is used to provide the source-domain prototypes $P_R$ required for the Source Structure Preservation (SSP) constraint. We adopt an offline pre-computation strategy for the reference detector $G_R$. Instead of maintaining a copy of $G_R$ in the GPU memory during the training of the detector $G$, we pre-calculate the prototype matrices for all source images and cache them on the disk. Since only matrices are saved, the storage cost is negligible.

For the IF-CARLA benchmark, the learning rate is set to 5e-4 with a batch size of 2. For the IF-CCT benchmark, we use a learning rate of 1e-4 and a batch size of 2. For the IF-LUNA16 benchmark, the learning rate is 1e-3 and the batch size was 4. The total training iterations for IF-CARLA and IF-CCT are 25k, while for IF-LUNA16 is 30k. The learning rate is decayed by a factor of 0.1 at the last 5k iterations for all benchmarks. We adopt the default SGD optimizer configuration provided by Detectron2. We follow the data loading and augmentation strategy proposed in AT~\cite{at} for our method and all compared baselines.

\begin{table}[t!]
\caption{The statistical details of the IF-CARLA benchmark.}
\label{tab:supp_if_carla}
\centering
\setlength{\tabcolsep}{6pt}
\begin{tabular}{cc|c|cc} 
\hline
\multicolumn{2}{c|}{\multirow{2}{*}{Subset}} & \multicolumn{1}{c|}{Source} & \multicolumn{2}{c}{Target} \\ 
\cline{3-5} 
\multicolumn{2}{c|}{} & Train & Train & Val \\ \hline
\multicolumn{2}{c|}{Sample}                            & 2,006         & 1,820         & 604         \\ \hline
\multicolumn{1}{c|}{\multirow{3}{*}{Object}} & Person  & 12,394        & -             & 4,701       \\
\multicolumn{1}{c|}{}                        & Vehicle & 12,207        & -             & 4,004       \\ \cline{2-5} 
\multicolumn{1}{c|}{}                        & Total   & 24,601        & -             & 8,705       \\ \hline
\end{tabular}
\end{table}

\begin{table}[t!]
\caption{The statistical details of the IF-CCT benchmark.}
\label{tab:supp_if_cct}
\centering
\setlength{\tabcolsep}{6pt}
\begin{tabular}{cc|c|cc} 
\hline
\multicolumn{2}{c|}{\multirow{2}{*}{Subset}} & \multicolumn{1}{c|}{Source} & \multicolumn{2}{c}{Target} \\ 
\cline{3-5} 
\multicolumn{2}{c|}{} & Train & Train & Val \\ \hline
\multicolumn{2}{c|}{Sample}                           & 2,872         & 2,678         & 534        \\ \hline
\multicolumn{1}{c|}{\multirow{6}{*}{Object}} & Rabbit & 511          & -            & 94         \\
\multicolumn{1}{c|}{}                        & Deer   & 648          & -            & 96         \\
\multicolumn{1}{c|}{}                        & Coyote & 734          & -            & 99         \\
\multicolumn{1}{c|}{}                        & Bobcat & 401          & -            & 93         \\
\multicolumn{1}{c|}{}                        & Bird   & 863          & -            & 99         \\ \cline{2-5} 
\multicolumn{1}{c|}{}                        & Total  & 3,157         & -            & 481        \\ \hline
\end{tabular}
\end{table}

\begin{table}[t!]
\caption{The statistical details of the IF-LUNA16 benchmark.}
\label{tab:supp_if_luna16}
\centering
\setlength{\tabcolsep}{6pt}
\begin{tabular}{cc|c|cc} 
\hline
\multicolumn{2}{c|}{\multirow{2}{*}{Subset}} & \multicolumn{1}{c|}{Source} & \multicolumn{2}{c}{Target} \\ 
\cline{3-5} 
\multicolumn{2}{c|}{} & Train & Train & Val \\ \hline
\multicolumn{2}{c|}{Sample}          & 2,008         & 865          & 231        \\ \hline
\multicolumn{1}{l|}{Object} & Nodule & 2,401         & -            & 258        \\ \hline
\end{tabular}
\end{table}

\section{More Benchmark Information}
\label{sec:supp_more_benchmark_information}
\begin{figure*}[]
  \centering
   \includegraphics[width=1.0\linewidth]{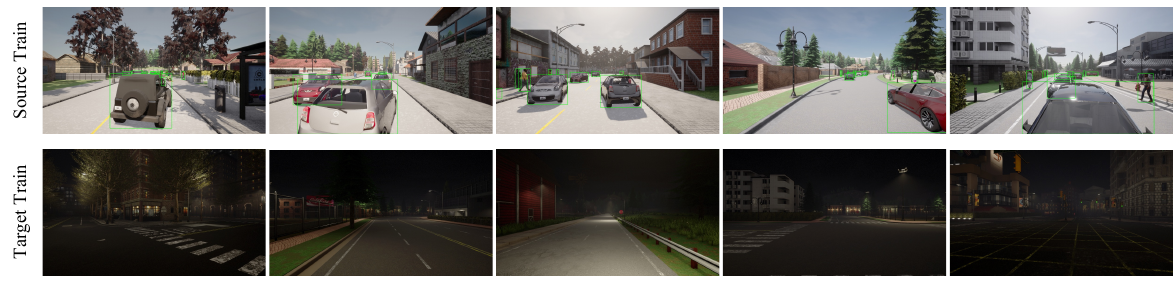}
   \caption{More training samples of the IF-CARLA benchmark.}
   \label{fig:A4_CARLA}
\end{figure*}

\begin{figure*}[]
  \centering
   \includegraphics[width=1.0\linewidth]{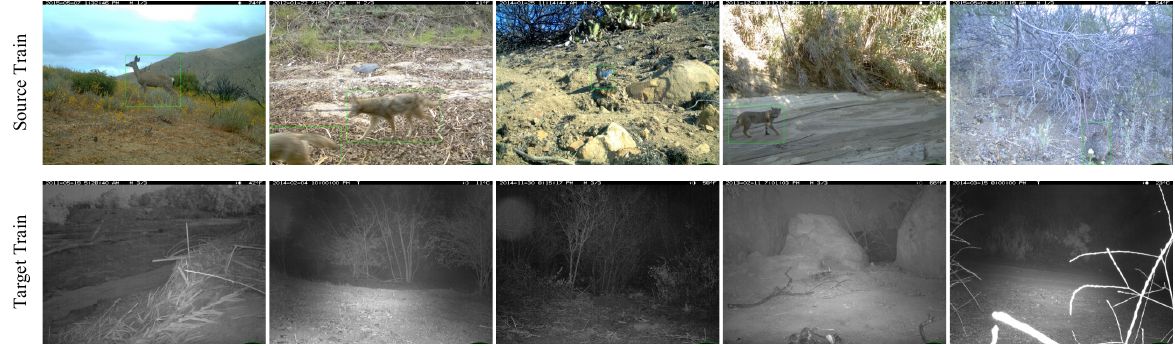}
   \caption{More training samples of the IF-CCT benchmark.}
   \label{fig:A5_CCT}
\end{figure*}

\begin{figure*}[]
  \centering
   \includegraphics[width=1.0\linewidth]{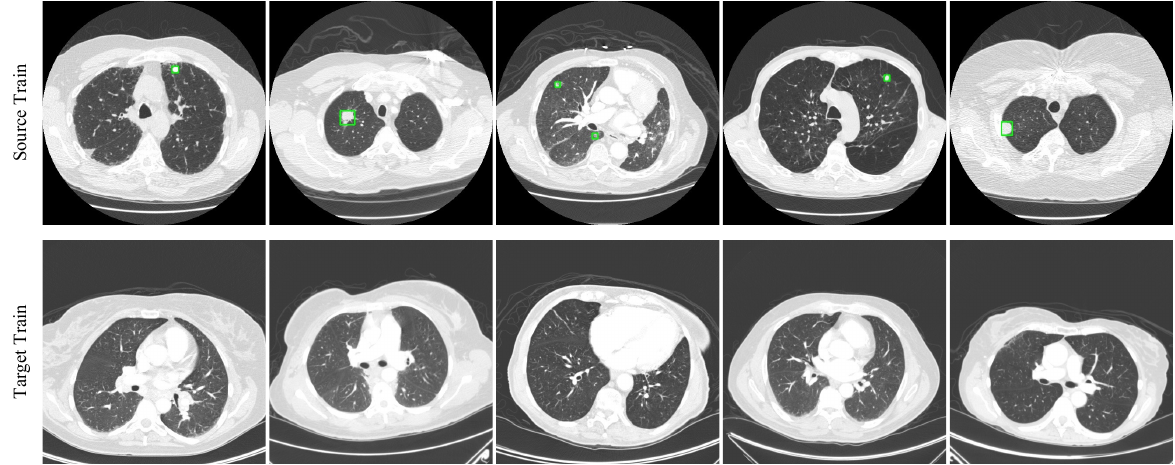}
   \caption{More training samples of the IF-LUNA16 benchmark. Zoom in for better viewing.}
   \label{fig:A6_LUNA16}
\end{figure*}

We established three dedicated benchmarks covering driving simulation, ecological monitoring, and medical diagnosis. Specifically, the IF-CARLA benchmark offers a simulated autonomous driving scenario with a day-to-night domain shift. It provides a strictly controlled environment with precise annotations, serving as a reliable foundation to support future research. The IF-CCT benchmark simulates a cross-modality transfer (visible light to infrared) for wildlife detection. This benchmark highlights the practical necessity of Instance-Free DAOD in the wild, where capturing positive animal instances is prohibitively costly and rare, while empty background frames are abundant. Finally, the IF-LUNA16 benchmark targets cross-device adaptation for pulmonary nodule detection. It reflects the clinical reality where varied scanning devices create domain gaps, and positive pathological cases are scarce compared to the easily accessible healthy screenings on newly deployed scanners. More visualizations of the source and target training samples are shown respectively in \cref{fig:A4_CARLA}, \cref{fig:A5_CCT} and \cref{fig:A6_LUNA16}.

Detailed statistic specifications for the IF-CARLA, IF-CCT, and IF-LUNA16 benchmarks are provided in \cref{tab:supp_if_carla}, \cref{tab:supp_if_cct}, and \cref{tab:supp_if_luna16}, respectively. We use the version 0.9.15 of CARLA~\cite{carla} to build the IF-CARLA. As introduced in \cref{sec:benchmarks}, the source-domain training set, the target-domain training set, and the target-domain validation set of IF-CARLA are split from the total of 4,430 camera viewpoints. Specifically, the source-domain training set is sampled from Town01, Town02 and Town04. The target-domain training set is sampled from Town05, Town07, and Town10. The target-domain validation set is sampled from Town03. The full triplet dataset containing 13,290 images (three samples for each viewpoint) will be also released.

We applied rigorous data processing procedures to ensure the high quality of the annotations. For the IF-CARLA benchmark, we deployed an RGB camera and a corresponding instance segmentation camera with identical intrinsic and extrinsic parameters at each viewpoint. By converting the generated instance segmentation masks into bounding box coordinates, we obtained pixel-perfect annotations. For the IF-LUNA16 benchmark, the original dataset provides annotations only as spatial centroids and radius. However, given the irregular shapes of pulmonary nodules, directly projecting these spherical parameters onto 2D slices often results in misalignment. To address this, we performed manual correction on the projected annotations to ensure precise bounding boxes for all nodule instances.


\section{Detection Results Visualization}
\label{sec:supp_detection_results_visualization}
\begin{figure*}[]
  \centering
   \includegraphics[width=1.0\linewidth]{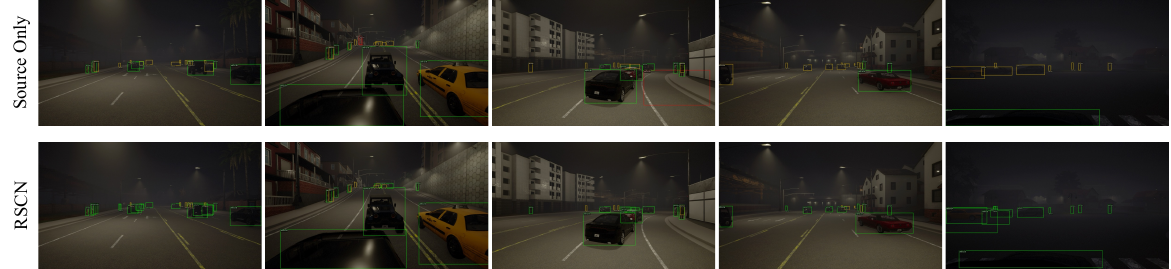}
   \caption{The visualizations of the detection results on the IF-CARLA benchmark. Correctly detected objects (True Positives, TP) are indicated by green bounding boxes. Incorrectly detected objects (False Positives, FP) are indicated by red bounding boxes, and missed ground truth objects (False Negatives, FN) are indicated by yellow bounding boxes.}
   \label{fig:A1_CARLA}
\end{figure*}

\begin{figure*}[]
  \centering
   \includegraphics[width=1.0\linewidth]{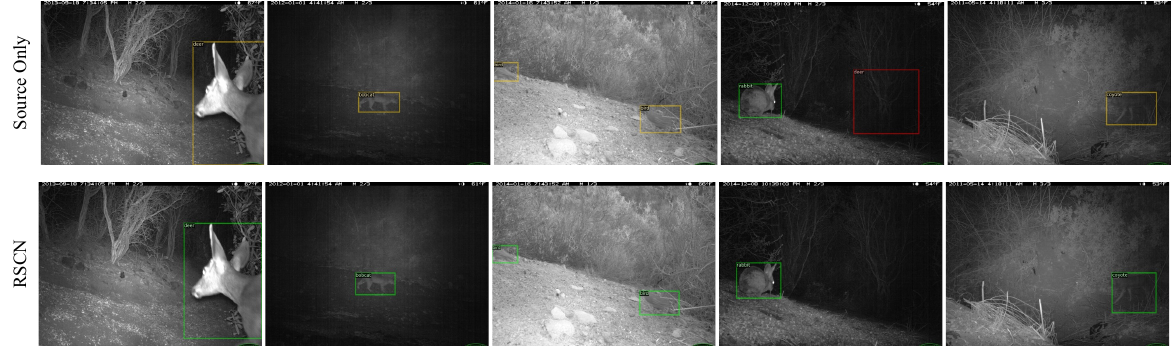}
   \caption{The visualizations of the detection results on the IF-CCT benchmark. Correctly detected objects (True Positives, TP) are indicated by green bounding boxes. Incorrectly detected objects (False Positives, FP) are indicated by red bounding boxes, and missed ground truth objects (False Negatives, FN) are indicated by yellow bounding boxes. }
   \label{fig:A2_CCT}
\end{figure*}

\begin{figure*}[]
  \centering
   \includegraphics[width=1.0\linewidth]{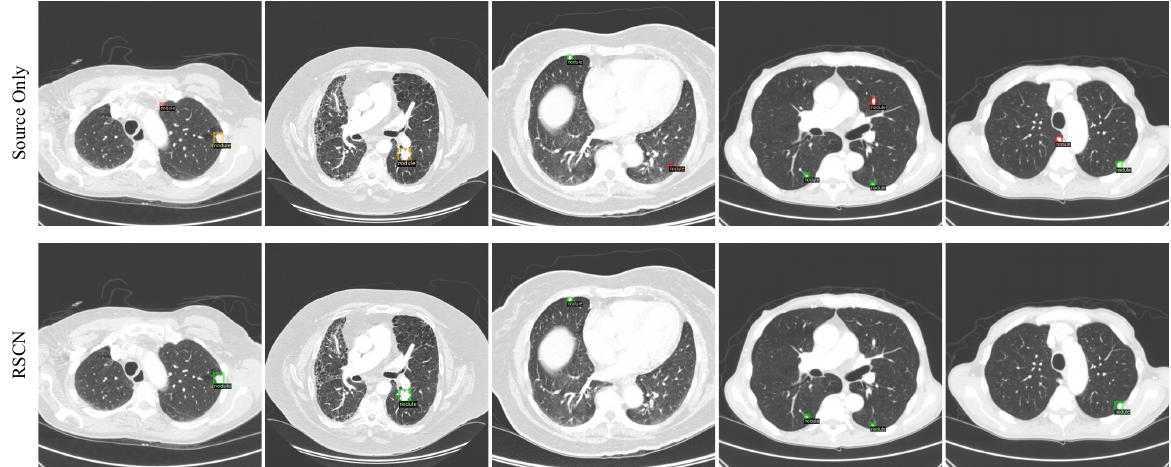}
   \caption{The visualizations of the detection results on the IF-LUNA16 benchmark. Correctly detected objects (True Positives, TP) are indicated by green bounding boxes. Incorrectly detected objects (False Positives, FP) are indicated by red bounding boxes, and missed ground truth objects (False Negatives, FN) are indicated by yellow bounding boxes.}
   \label{fig:A3_LUNA16}
\end{figure*}

In this section, we provide visualizations of the detection results, comparing the performance of our \methodname against the source-only baseline. The visualizations of the IF-CARLA, IF-CCT, and IF-LUNA16 benchmarks are shown in \cref{fig:A1_CARLA}, \cref{fig:A2_CCT} and \cref{fig:A3_LUNA16} respectively. As illustrated in the second and third columns of \cref{fig:A1_CARLA}, the fourth column of \cref{fig:A2_CCT}, and the first, third, fourth and fifth columns of \cref{fig:A3_LUNA16}, the Source Only model suffers from false positives (FPs) due to the domain shifts. Furthermore, these shifts leads to numerous missed detections (False Negatives, FNs), for example, all columns of the \cref{fig:A1_CARLA}, the first, second, third, and fifth columns of \cref{fig:A2_CCT}, and the first and second columns of \cref{fig:A3_LUNA16}. In contrast, our proposed \methodname effectively mitigates these issues, reducing FPs and recovering the challenging FN objects. These results demonstrate that our proposed method achieves superior transfer performance, effectively detecting foreground objects in the target domain despite their absence during training.

\section{Future Work Discussion}
\label{sec:supp_future_work_discussion}
In this work, we addressed the Instance-Free DAOD challenge by proposing the \methodname, which leverages background prototype alignment and relative geometric consistency to bridge the domain gap without observing target foregrounds. We believe this novel Instance-Free DAOD setting opens up more avenues for future research.
One promising direction is the explicit construction or synthesis of target-domain foreground representations. Since the difficulty of Instance-Free DAOD lies in the absence of real target instances, future works could explore feature disentanglement techniques to address this deficit. By disentangling feature representation into domain-invariant content (\eg, object semantics) and domain-specific style (\eg, background texture or illumination), it may be possible to recombine source-domain content with target-domain style. This would allow for the generation of hallucinated target-like foreground features, turning the unsupervised problem into a supervised one with synthesized data. We hope our benchmark and baseline can serve as a foundation for further explorations.

%% file: example_paper.bib
@String(CVPR= {IEEE Conf. Comput. Vis. Pattern Recog.})

@String(ICCV= {Int. Conf. Comput. Vis.})

@String(ECCV= {Eur. Conf. Comput. Vis.})

@String(ICLR = {Int. Conf. Learn. Represent.})

@String(AAAI = {AAAI})

@String(CVPR  = {CVPR})

@String(ICCV  = {ICCV})

@String(ECCV  = {ECCV})

@String(ICLR  = {ICLR})

@inproceedings{resnet,
  title={Deep residual learning for image recognition},
  author={He, Kaiming and Zhang, Xiangyu and Ren, Shaoqing and Sun, Jian},
  booktitle={CVPR},
  pages={770--778},
  year={2016}
}

@inproceedings{vgg,
  title={Very deep convolutional networks for large-scale image recognition},
  author={Simonyan, K and Zisserman, A},
  booktitle={ICLR},
  year={2015},
  organization={Computational and Biological Learning Society}
}

@article{transformer,
  title={Attention is all you need},
  author={Vaswani, Ashish and Shazeer, Noam and Parmar, Niki and Uszkoreit, Jakob and Jones, Llion and Gomez, Aidan N and Kaiser, {\L}ukasz and Polosukhin, Illia},
  journal={NeurIPS},
  volume={30},
  year={2017}
}

@article{lenet,
  title={Imagenet classification with deep convolutional neural networks},
  author={Krizhevsky, Alex and Sutskever, Ilya and Hinton, Geoffrey E},
  journal={NeurIPS},
  volume={25},
  year={2012}
}

@inproceedings{rcnn,
  title={Rich feature hierarchies for accurate object detection and semantic segmentation},
  author={Girshick, Ross and Donahue, Jeff and Darrell, Trevor and Malik, Jitendra},
  booktitle={CVPR},
  pages={580--587},
  year={2014}
}

@inproceedings{fastrcnn,
  title={Fast r-cnn},
  author={Girshick, Ross},
  booktitle={ICCV},
  pages={1440--1448},
  year={2015}
}

@article{fasterrcnn,
  title={Faster r-cnn: Towards real-time object detection with region proposal networks},
  author={Ren, Shaoqing and He, Kaiming and Girshick, Ross and Sun, Jian},
  journal={NeurIPS},
  volume={28},
  year={2015}
}

@inproceedings{yolo,
  title={You only look once: Unified, real-time object detection},
  author={Redmon, Joseph and Divvala, Santosh and Girshick, Ross and Farhadi, Ali},
  booktitle={CVPR},
  pages={779--788},
  year={2016}
}

@inproceedings{retinanet,
  title={Focal loss for dense object detection},
  author={Lin, Tsung-Yi and Goyal, Priya and Girshick, Ross and He, Kaiming and Doll{\'a}r, Piotr},
  booktitle={ICCV},
  pages={2980--2988},
  year={2017}
}

@inproceedings{detr,
  title={End-to-end object detection with transformers},
  author={Carion, Nicolas and Massa, Francisco and Synnaeve, Gabriel and Usunier, Nicolas and Kirillov, Alexander and Zagoruyko, Sergey},
  booktitle={ECCV},
  pages={213--229},
  year={2020},
  organization={Springer}
}

@article{deformabledetr,
  title={Deformable detr: Deformable transformers for end-to-end object detection},
  author={Zhu, Xizhou and Su, Weijie and Lu, Lewei and Li, Bin and Wang, Xiaogang and Dai, Jifeng},
  journal={arXiv preprint arXiv:2010.04159},
  year={2020}
}

@article{dino,
  title={Dino: Detr with improved denoising anchor boxes for end-to-end object detection},
  author={Zhang, Hao and Li, Feng and Liu, Shilong and Zhang, Lei and Su, Hang and Zhu, Jun and Ni, Lionel M and Shum, Heung-Yeung},
  journal={arXiv preprint arXiv:2203.03605},
  year={2022}
}

@article{dann,
  title={Domain-adversarial training of neural networks},
  author={Ganin, Yaroslav and Ustinova, Evgeniya and Ajakan, Hana and Germain, Pascal and Larochelle, Hugo and Laviolette, Fran{\c{c}}ois and March, Mario and Lempitsky, Victor},
  journal={JMLR},
  volume={17},
  number={59},
  pages={1--35},
  year={2016}
}

@article{theory,
  title={A theory of learning from different domains},
  author={Ben-David, Shai and Blitzer, John and Crammer, Koby and Kulesza, Alex and Pereira, Fernando and Vaughan, Jennifer Wortman},
  journal={Machine learning},
  volume={79},
  number={1},
  pages={151--175},
  year={2010},
  publisher={Springer}
}

@inproceedings{daf,
  title={Domain adaptive faster r-cnn for object detection in the wild},
  author={Chen, Yuhua and Li, Wen and Sakaridis, Christos and Dai, Dengxin and Van Gool, Luc},
  booktitle={CVPR},
  pages={3339--3348},
  year={2018}
}

@inproceedings{swda,
  title={Strong-weak distribution alignment for adaptive object detection},
  author={Saito, Kuniaki and Ushiku, Yoshitaka and Harada, Tatsuya and Saenko, Kate},
  booktitle={CVPR},
  pages={6956--6965},
  year={2019}
}

@inproceedings{divmatch,
  title={Diversify and match: A domain adaptive representation learning paradigm for object detection},
  author={Kim, Taekyung and Jeong, Minki and Kim, Seunghyeon and Choi, Seokeon and Kim, Changick},
  booktitle={CVPR},
  pages={12456--12465},
  year={2019}
}

@inproceedings{scda,
  title={Adapting object detectors via selective cross-domain alignment},
  author={Zhu, Xinge and Pang, Jiangmiao and Yang, Ceyuan and Shi, Jianping and Lin, Dahua},
  booktitle={CVPR},
  pages={687--696},
  year={2019}
}

@inproceedings{htcn,
  title={Harmonizing transferability and discriminability for adapting object detectors},
  author={Chen, Chaoqi and Zheng, Zebiao and Ding, Xinghao and Huang, Yue and Dou, Qi},
  booktitle={CVPR},
  pages={8869--8878},
  year={2020}
}

@inproceedings{everypixelmatters,
  title={Every pixel matters: Center-aware feature alignment for domain adaptive object detector},
  author={Hsu, Cheng-Chun and Tsai, Yi-Hsuan and Lin, Yen-Yu and Yang, Ming-Hsuan},
  booktitle={ECCV},
  pages={733--748},
  year={2020},
  organization={Springer}
}

@inproceedings{megacda,
  title={Mega-cda: Memory guided attention for category-aware unsupervised domain adaptive object detection},
  author={Vs, Vibashan and Gupta, Vikram and Oza, Poojan and Sindagi, Vishwanath A and Patel, Vishal M},
  booktitle={CVPR},
  pages={4516--4526},
  year={2021}
}

@inproceedings{sigma,
  title={Sigma: Semantic-complete graph matching for domain adaptive object detection},
  author={Li, Wuyang and Liu, Xinyu and Yuan, Yixuan},
  booktitle={CVPR},
  pages={5291--5300},
  year={2022}
}

@article{sigma++,
  title={Sigma++: Improved semantic-complete graph matching for domain adaptive object detection},
  author={Li, Wuyang and Liu, Xinyu and Yuan, Yixuan},
  journal={T-PAMI},
  volume={45},
  number={7},
  pages={9022--9040},
  year={2023},
  publisher={IEEE}
}

@inproceedings{umt,
  title={Unbiased mean teacher for cross-domain object detection},
  author={Deng, Jinhong and Li, Wen and Chen, Yuhua and Duan, Lixin},
  booktitle={CVPR},
  pages={4091--4101},
  year={2021}
}

@inproceedings{at,
  title={Cross-domain adaptive teacher for object detection},
  author={Li, Yu-Jhe and Dai, Xiaoliang and Ma, Chih-Yao and Liu, Yen-Cheng and Chen, Kan and Wu, Bichen and He, Zijian and Kitani, Kris and Vajda, Peter},
  booktitle={CVPR},
  pages={7581--7590},
  year={2022}
}

@inproceedings{ht,
  title={Harmonious teacher for cross-domain object detection},
  author={Deng, Jinhong and Xu, Dongli and Li, Wen and Duan, Lixin},
  booktitle={CVPR},
  pages={23829--23838},
  year={2023}
}

@article{ssal,
  title={Ssal: Synergizing between self-training and adversarial learning for domain adaptive object detection},
  author={Munir, Muhammad Akhtar and Khan, Muhammad Haris and Sarfraz, M and Ali, Mohsen},
  journal={NeurIPS},
  volume={34},
  pages={22770--22782},
  year={2021}
}

@inproceedings{pt,
  title={Learning Domain Adaptive Object Detection with Probabilistic Teacher},
  author={Chen, Meilin and Chen, Weijie and Yang, Shicai and Song, Jie and Wang, Xinchao and Zhang, Lei and Yan, Yunfeng and Qi, Donglian and Zhuang, Yueting and Xie, Di and others},
  booktitle={ICML},
  pages={3040--3055},
  year={2022},
  organization={PMLR}
}

@inproceedings{mttrans,
  title={MTTrans: Cross-domain object detection with mean teacher transformer},
  author={Yu, Jinze and Liu, Jiaming and Wei, Xiaobao and Zhou, Haoyi and Nakata, Yohei and Gudovskiy, Denis and Okuno, Tomoyuki and Li, Jianxin and Keutzer, Kurt and Zhang, Shanghang},
  booktitle={ECCV},
  pages={629--645},
  year={2022},
  organization={Springer}
}

@inproceedings{cmt,
  title={Contrastive mean teacher for domain adaptive object detectors},
  author={Cao, Shengcao and Joshi, Dhiraj and Gui, Liang-Yan and Wang, Yu-Xiong},
  booktitle={CVPR},
  pages={23839--23848},
  year={2023}
}

@inproceedings{cat,
  title={Cat: Exploiting inter-class dynamics for domain adaptive object detection},
  author={Kennerley, Mikhail and Wang, Jian-Gang and Veeravalli, Bharadwaj and Tan, Robby T},
  booktitle={CVPR},
  pages={16541--16550},
  year={2024}
}

@inproceedings{cct,
  title={Recognition in terra incognita},
  author={Beery, Sara and Van Horn, Grant and Perona, Pietro},
  booktitle={ECCV},
  pages={456--473},
  year={2018}
}

@article{luna16,
  title={Validation, comparison, and combination of algorithms for automatic detection of pulmonary nodules in computed tomography images: the LUNA16 challenge},
  author={Setio, Arnaud Arindra Adiyoso and Traverso, Alberto and De Bel, Thomas and Berens, Moira SN and Van Den Bogaard, Cas and Cerello, Piergiorgio and Chen, Hao and Dou, Qi and Fantacci, Maria Evelina and Geurts, Bram and others},
  journal={Medical Image Analysis},
  volume={42},
  pages={1--13},
  year={2017},
  publisher={Elsevier}
}

@inproceedings{rescue,
  title={Nomad: A natural, occluded, multi-scale aerial dataset, for emergency response scenarios},
  author={Russell Bernal, Arturo Miguel and Scheirer, Walter and Cleland-Huang, Jane},
  booktitle={WACV},
  pages={8584--8595},
  year={2024}
}

@inproceedings{ssd,
  title={Ssd: Single shot multibox detector},
  author={Liu, Wei and Anguelov, Dragomir and Erhan, Dumitru and Szegedy, Christian and Reed, Scott and Fu, Cheng-Yang and Berg, Alexander C},
  booktitle={ECCV},
  pages={21--37},
  year={2016},
  organization={Springer}
}

@inproceedings{maskrcnn,
  title={Mask r-cnn},
  author={He, Kaiming and Gkioxari, Georgia and Doll{\'a}r, Piotr and Girshick, Ross},
  booktitle={Proceedings of the IEEE international conference on computer vision},
  pages={2961--2969},
  year={2017}
}

@inproceedings{cascadercnn,
  title={Cascade r-cnn: Delving into high quality object detection},
  author={Cai, Zhaowei and Vasconcelos, Nuno},
  booktitle={Proceedings of the IEEE conference on computer vision and pattern recognition},
  pages={6154--6162},
  year={2018}
}

@inproceedings{fcos,
  title={Fcos: Fully convolutional one-stage object detection},
  author={Tian, Zhi and Shen, Chunhua and Chen, Hao and He, Tong},
  booktitle={ICCV},
  pages={9627--9636},
  year={2019}
}

@article{centernet,
  title={Objects as points},
  author={Zhou, Xingyi and Wang, Dequan and Kr{\"a}henb{\"u}hl, Philipp},
  journal={arXiv preprint arXiv:1904.07850},
  year={2019}
}

@inproceedings{multi,
  title={Multi-adversarial faster-rcnn for unrestricted object detection},
  author={He, Zhenwei and Zhang, Lei},
  booktitle={ICCV},
  pages={6668--6677},
  year={2019}
}

@article{mt,
  title={Mean teachers are better role models: Weight-averaged consistency targets improve semi-supervised deep learning results},
  author={Tarvainen, Antti and Valpola, Harri},
  journal={NeurIPS},
  volume={30},
  year={2017}
}

@inproceedings{collapse,
  title={Beware of model collapse! fast and stable test-time adaptation for robust question answering},
  author={Su, Yi and Ji, Yixin and Li, Juntao and Ye, Hai and Zhang, Min},
  booktitle={EMNLP},
  pages={12998--13011},
  year={2023}
}

@article{collapse2,
  title={Model adaptation: Historical contrastive learning for unsupervised domain adaptation without source data},
  author={Huang, Jiaxing and Guan, Dayan and Xiao, Aoran and Lu, Shijian},
  journal={NeurIPS},
  volume={34},
  pages={3635--3649},
  year={2021}
}

@article{lidc-ldri,
  title={The lung image database consortium (LIDC) and image database resource initiative (IDRI): a completed reference database of lung nodules on CT scans},
  author={Armato III, Samuel G and McLennan, Geoffrey and Bidaut, Luc and McNitt-Gray, Michael F and Meyer, Charles R and Reeves, Anthony P and Zhao, Binsheng and Aberle, Denise R and Henschke, Claudia I and Hoffman, Eric A and others},
  journal={Medical physics},
  volume={38},
  number={2},
  pages={915--931},
  year={2011},
  publisher={Wiley Online Library}
}

@inproceedings{carla,
  title={CARLA: An open urban driving simulator},
  author={Dosovitskiy, Alexey and Ros, German and Codevilla, Felipe and Lopez, Antonio and Koltun, Vladlen},
  booktitle={CoRL},
  pages={1--16},
  year={2017},
  organization={PMLR}
}

@inproceedings{mad,
  title={Multi-view adversarial discriminator: Mine the non-causal factors for object detection in unseen domains},
  author={Xu, Mingjun and Qin, Lingyun and Chen, Weijie and Pu, Shiliang and Zhang, Lei},
  booktitle={CVPR},
  pages={8103--8112},
  year={2023}
}

@inproceedings{vdd,
  title={Vector-decomposed disentanglement for domain-invariant object detection},
  author={Wu, Aming and Liu, Rui and Han, Yahong and Zhu, Linchao and Yang, Yi},
  booktitle={ICCV},
  pages={9342--9351},
  year={2021}
}

@inproceedings{he2025differential,
  title={Differential Alignment for Domain Adaptive Object Detection},
  author={He, Xinyu and Li, Xinhui and Guo, Xiaojie},
  booktitle={AAAI},
  volume={39},
  number={16},
  pages={17150--17158},
  year={2025}
}

@inproceedings{cui2025debiased,
  title={Debiased Teacher for Day-to-Night Domain Adaptive Object Detection},
  author={Cui, Yiming and Li, Liang and Yin, Haibing and Gao, Yuhan and Sun, Yaoqi and Yan, Chenggang},
  booktitle={ICCV},
  pages={2577--2587},
  year={2025}
}

@inproceedings{cyclegan,
  title={Unpaired image-to-image translation using cycle-consistent adversarial networks},
  author={Zhu, Jun-Yan and Park, Taesung and Isola, Phillip and Efros, Alexei A},
  booktitle={ICCV},
  pages={2223--2232},
  year={2017}
}
